\documentclass[conference]{IEEEtran}
\IEEEoverridecommandlockouts
\usepackage{cite}
\usepackage{amsmath,amssymb,amsfonts}
\usepackage{algorithmic}
\usepackage[ruled,vlined]{algorithm2e}
\usepackage{algorithmic}
\usepackage{ifthen}
\usepackage{graphicx}
\usepackage{textcomp}
\usepackage{xcolor}
\usepackage{xspace}
\usepackage{subcaption}
\usepackage{graphicx}
\usepackage{lipsum}
\usepackage{url}
\DeclareCaptionFont{mysize}{\fontsize{8}{9.6}\selectfont}
\usepackage[subtle,tracking=normal]{savetrees}
\def\BibTeX{{\rm B\kern-.05em{\sc i\kern-.025em b}\kern-.08em
    T\kern-.1667em\lower.7ex\hbox{E}\kern-.125emX}}
\begin{document}

\title{Surrogate-Assisted Genetic Algorithm\\for Wrapper Feature Selection}

\author{
    \IEEEauthorblockN{Mohammed Ghaith Altarabichi, S\l{}awomir Nowaczyk,
   Sepideh Pashami\\
    and Peyman Sheikholharam Mashhadi}
    \IEEEauthorblockA{\textit{Center for Applied Intelligent Systems Research, Halmstad University, Sweden}\\}
    \IEEEauthorblockA{\{mohammed\_ghaith.altarabichi, slawomir.nowaczyk, sepideh.pashami, peyman.mashhadi\}@hh.se}
    \thanks{© 2021 IEEE.  Personal use of this material is permitted.  Permission from IEEE must be obtained for all other uses, in any current or future media, including reprinting/republishing this material for advertising or promotional purposes, creating new collective works, for resale or redistribution to servers or lists, or reuse of any copyrighted component of this work in other works.}
}


\newcommand{\SAGA}{\emph{SAGA}\xspace}
\newcommand{\SAGAso}{\emph{SAGA[so=1]}\xspace}

\IEEEoverridecommandlockouts

\IEEEpubid{\makebox[\columnwidth]{978-1-7281-8393-0/21/\$31.00~\copyright2021 IEEE \hfill}
\hspace{\columnsep}\makebox[\columnwidth]{ }}

\maketitle

\IEEEpubidadjcol

\begin{abstract}
Feature selection is an intractable problem, therefore practical algorithms often trade off the solution accuracy against the computation time. In this paper, we propose a novel multi-stage feature selection framework utilizing multiple levels of approximations, or surrogates. Such a framework allows for using wrapper approaches in a much more computationally efficient way, significantly increasing the quality of feature selection solutions achievable, especially on large datasets. We design and evaluate a Surrogate-Assisted Genetic Algorithm $(SAGA)$ which utilizes this concept to guide the evolutionary search during the early phase of exploration. $SAGA$ only switches to evaluating the original function at the final exploitation phase.

We prove that the run-time upper bound of $SAGA$ surrogate-assisted stage is at worse equal to the wrapper $GA$, and it scales better for induction algorithms of high order of complexity in number of instances. We demonstrate, using 14 datasets from the UCI ML repository, that in practice $SAGA$ significantly reduces the computation time compared to a baseline wrapper Genetic Algorithm $(GA)$, while converging to solutions of significantly higher accuracy. Our experiments show that $SAGA$ can arrive at near-optimal solutions three times faster than a wrapper $GA$, on average. We also showcase the importance of evolution control approach designed to prevent surrogates from misleading the evolutionary search towards false optima. 
\end{abstract}


\section{Introduction}
Feature Selection is an important data mining technique used to identify the most informative subsets of features for a given learning task. Primarily, feature selection aims to achieve two main goals: (a) reduce the computational complexity of model training by using fewer features; (b) improve generalization performance and model accuracy by reducing the overfitting. The problem of finding the best feature subset is known to be NP-hard \cite{np}. Therefore, feature selection is often performed in a greedy manner to reduce computation time. The performance of a greedy approach is shown to be inferior to a global search \cite{greedy1}, in particular for high-dimensional datasets with many interacting features \cite{greedy2}. In this paper, we propose a novel feature selection framework that utilizes surrogates to reduce the computation time of a wrapper approach, while at the same time improving the accuracy of discovered solutions. 


Our algorithm $(SAGA)$, Surrogate-Assisted Genetic Algorithm, is based on concepts derived from the research on progressive sampling and optimization with approximate fitness function. $SAGA$ utilizes multiple levels of surrogates to allow the evolutionary search for computationally-efficient exploration early on, and highly-accurate exploitation in the final stages. A surrogate can be any approximation of the true objective function within the wrapper model, and in our work, it refers to a machine learning model trained with a sub-sample of training instances. $SAGA$ starts by constructing the initial surrogate using a small batch of instances, and explores it using a large population $GA$. Once the usefulness of that is exhausted, it switches to the next level, with a new surrogate. That new surrogate is initialized with the best found solutions, and allowed to access a larger set of data instances. Meanwhile, the $GA$ population is reduced to keep the computation time manageable. This process continues iteratively, as the surrogates move progressively towards more accurate approximations, while the $GA$ population size keeps decreasing. Finally, the complete dataset is used in the ultimate exploitation stage, focusing on the most promising solution neighborhood.

In other words, the proposed framework divides the evolutionary search into multiple phases, progressively lowering the focus on exploration and increasing the focus on exploitation. We experimentally show that surrogates can arrive at near-optimal solutions in a fraction of the time required by the original function. Therefore, $SAGA$ utilizes computationally cheap surrogates to guide the evolutionary search during early exploration. However, ultimately the surrogates end up having too low fidelity, which is a critical impediment in terms of exploitation. Therefore, $SAGA$ switches to the original function at the very end, for full exploitation, once the surrogate-stage has reached its limit.


We show that the theoretical run-time upper bound of $SAGA$ surrogate-assisted stage is equal to that of a wrapper $GA$. Moreover, $SAGA$ is more scalable for induction algorithms of high order polynomial complexity in number of instances. Our algorithm is shown to outperform a wrapper $GA$ by converging to solutions of significantly higher accuracy in a significantly shorter time. We also demonstrate that the proposed evolution control saves time significantly while offering higher average accuracy during the exploration phase.



\section{Related Work}
A recent survey on evolutionary computation approaches of feature selection \cite{evolutionary-fs-survey} recognized computational cost as a major research challenge in the field. The survey labeled existing feature selection $GA$ approaches as "unfit to solve big data tasks", as they found majority of research to be limited to small datasets, i.e., less than 1\,000 instances.

The same survey classified feature selection $GA$ approaches into three main categories: filter, wrapper, and hybrid. A Filter approach utilizes measures based on inconsistency rate \cite{filter-consistency}, information theory \cite{filter-information}, fuzzy set theory \cite{filter-fuzzy}, and rough set theory \cite{filter-rough} to evaluate fitness of feature subsets. A filter approach is usually faster than a wrapper approach \cite{filter-consistency}. However, filter methods share the fundamental limitation of being agnostic towards the induction algorithm \cite{agnostic}. 

The hybrid approaches \cite{hybrid} trade off accuracy for computational efficiency by first reducing the feature search space using a filter. They only apply a wrapper to the remaining selected features \cite{hybrid1,hybrid2,hybrid3,hybrid4,hybrid5}. The solution of this approach isn't scalable in terms of instances, as reduction is only applied to the feature search space. In addition, it ignores feature interactions, namely that low-ranked features might turn important when combined with other features.

The final approach is wrapper \cite{wrapper}, and $SAGA$ belongs to this category. Wrapper approaches rely on the induction algorithm to explicitly evaluate fitness of feature subsets. Therefore, a wrapper generally offers higher accuracy compared to filter, at the expense of computation time \cite{fs-survey, wrapper-better-1, wrapper-better-2}. An earlier work \cite{fast} incorporated approximate models to reduce the computation cost of the wrapper. They proposed ``training set sampling,'' method in which a light-weight k-Nearest Neighbors (kNN) model is used to approximate the computationally expensive fitness evaluations of a Neural Network. Additionally, only a portion of available data is used to train the $kNN$ model. However, the experiments were limited to small datasets. Also, re-sampling on each generation forced more evaluations, and consequently increased computational cost. 

The Optimal Sample Size $(OSS)$ \cite{ps}, defined as the smallest sample size that offers minimum achievable error for a given learning algorithm, is used by \cite{gaoss} to train a surrogate model. However, the algorithm offers no guarantee that a model trained with $OSS$ will maintain the quality of feature selection solutions comparing to using the complete dataset. 

 An approach based on the MapReduce paradigm is proposed in \cite{mapreduce}. The algorithm produces partial results in the map phase by decomposing the original dataset into blocks of instances, and merges results in the reduce phase. Although the work sufficiently motivates the choice of MapReduce paradigm to reduce the computational cost, the performance is only compared against baseline models trained with all the features, and only for two large datasets.

The mentioned wrapper methods share a common limitation. They never analyze how much they are compromising the accuracy to reduce computation time. They lack a theoretical or experimental evaluation against a wrapper trained with the same induction algorithm of the learning task, using all available training data. Additionally, there is no guarantee that a wrapper method based on a certain class of induction algorithms, i.e., kNN is good universally at selecting features for a task using a different class of induction algorithm i.e., Neural Network.  


In terms of evolutionary optimization, a recent overview \cite{recent-sa-survey} acknowledged the importance of incorporating surrogate-assisted evolutionary approaches to solve large scale problems. The same survey justified the limited success realized in this line of research by lack of benchmarks, access to real-world data, and computational resources required for evaluation.  

An earlier survey on fitness approximation \cite{sampling-survey} identified the challenge of maintaining the quality of solutions while reducing computation time as the major drawback in existing surrogate-assisted evolution methods. The lack of theoretical work to guarantee that a surrogate-assisted approach can converge faster than a classical approach is an additional challenge recognized by \cite{sa-survey}.

To summarize, wrapper is consensus gold-standard in terms of solution quality, however, its practical applicability is limited due to prohibitive computational costs. This work is an important milestone that is unique in terms of demonstrating improvement both for accuracy and time.

\section{Method}
We start this section by formalizing the feature selection problem and outlining a general description of our method. We highlight the key components of the proposed surrogate-assisted framework in the following subsections. 

In the feature selection problem, we are given a training set $D$ of labeled pairs of the dimensions $(N\times K)$, of which $N$ represents the number of instances, and $K$ is the number of features. The goal is to select a new subspace $(N\times L)$ (where $L \leq K$), while improving classification/regression performance (using machine learning algorithm, or model, $M$) over the one obtained with the original feature space. In the $GA$ wrapper setting, a machine-learning model $M$ is trained to evaluate candidate feature subsets during the evolutionary search. In our work, we refer to evaluations done using $M$ trained on all instances from $D$ as the true, or original, fitness function evaluations; and evaluations using $M$ trained on a subset of instances from $D$ as a surrogate, or approximation.

$SAGA$ follows an iterative procedure of uniformly sampling a number of instances without replacement from $D$ on each surrogate level, according to a predefined sampling schedule $S_g$. In our algorithm, $S_g$ is a geometric schedule given by the following simple equation: 

\begin{equation}
S_g= \frac{N}{a^i} = \{\frac{N}{a^b} , \frac{N}{a^{b-1}},...,\frac{N}{a}\}
\label{gs-eq}
\end{equation}

\noindent where $b$ is the number of levels, $i=b,b-1,...,2,1$ and $a$ is a constant. According to $S_g$, a starting partial training sample $D_1$ is created by uniformly sampling $(\frac{N}{a^b})$ instances from $D$, without replacement. The first level surrogate model $M_1$ is only allowed to access training instances of $D_1$. The model $M_1$ is used to evaluate feature subsets accuracies in a $GA$ wrapper settings during the first level. The first level $GA$ is initiated with a random population according to the initial population size hyper-parameter $(p_0)$. Each individual encodes a single feature subset as a binary string. An individual $g$ can be expressed as $g \in \{0, 1\}^K$, where $1$ indicates the selection of the corresponding feature in this individual's chromosome. The creation of individuals is done according to a fixed independent probability $(0.5)$ of each feature to be selected.

An evolution control is designed to govern switching from one level to the next. In our framework, it can be characterized as fixed, individual-based \cite{sampling-survey}, and following best strategy \cite{best}. Evolution control of $SAGA$ reevaluates the best individual found by surrogate $M_i$ using $M$, after every fixed number of $(z)$ generations. The algorithm only keeps using $M_i$ if the fitness, according to $M$ is still improving, otherwise, it switches to the next level surrogate $M_{i+1}$.

Each next level surrogate model $M_{i+1}$ is allowed to access a larger training set, $D_{i+1}$.
The population of the wrapper $GA$ is initialized with the best individual found in so far in the search. This is done to ensure that the algorithm progressively builds upon earlier results. However, the population size is reduced on each level according to the population reduction rate, $(pr)$ hyper-parameter.

This iterative procedure continues until the last surrogate level, using model $M_b$. Once the surrogate stage is over, $SAGA$ migrates the best found individual $(g^\prime)$ to the final population, and uses $g^\prime$ as a seed to create the rest of it. This is done by using the frequency of ones (included features) of $g^\prime$ to create other individuals with approximately similar number of included features to $g^\prime$. The objective is to focus the search on the most promising neighborhood discovered by surrogates.



We provide a pseudocode of $SAGA$ algorithm in Algorithm~\ref{alg:saga}. In the following subsections We discuss the key components of the framework in more detail.

\subsection{Evolutionary Algorithm} 

Our framework is built to accommodate any variation of $GA$ (or even more broadly, any iterative optimization procedure) to carry out the feature selection in the wrapper settings. In this paper, we have used the $CHC$ Genetic Algorithm (Cross generational elitist selection, Heterogeneous recombination and Cataclysmic mutation) \cite{chcname} as it is shown to be computationally efficient \cite{chc}.

$CHC$ is different from a traditional $GA$ in a number of ways. First, it always guarantees the survival of the best individual found so far. The crossover operator of $CHC$ employs a method for incest prevention; parents are only allowed to mate if their measured Hamming distance is larger than a predefined threshold. The threshold is adaptively reduced as population diversity drops. Finally, no mutation operator is used; instead, the population is reinitialized to introduce diversity whenever the search stagnates. The best individual is migrated to the new population to ensure that the history of the search is not lost, and the chromosome of the best individual is utilized as a template to re-seed the other individuals in the population. A pseudocode of $CHC$ algorithm is provided in Algorithm~\ref{alg:chc}.


\subsection{Sampling Procedure} 
Broadly speaking, instances to train the surrogates could be selected in two different ways. The first idea is active data sampling, in which training samples are selected with the objective of improving a predefined surrogate quality measure \cite{sampling-survey}. An example of this procedure is a statistical active selection of data using integrated squared bias \cite{active-sampling}. Other metrics of measuring surrogate quality include rank, and continuous correlation \cite{quality-measures}. 
  
A much simpler alternative is random sampling. $SAGA$ uses uniform sampling without replacement, as the running time is a major motivation of this work. All active sampling methods carry a computational burden, often quite significant one. On the other hand, our algorithm is designed to compensate for imperfect surrogates. $SAGA$ accepts that a surrogate constructed using a random sample is unlikely to offer a global approximation of the original function as good as one coming from active sampling. Instead of spending time identifying the most informative individual samples, however, the algorithm mitigates this by progressively switching to a next level surrogate with access to more data. 
  
 \subsection{Sampling Schedule} 
 


In this subsection, we prove that the theoretical upper bound of $SAGA$ run-time is no worse than a traditional wrapper $GA$. In order to show this, we compare the expected run-time of sequentially running all surrogate levels according to $S_g= \{\frac{N}{a^b}, \frac{N}{a^{b-1}},...,\frac{N}{a}\}$, against a traditional wrapper running the simple schedule $S_c = \{ N\}$.  




Geometric sampling is shown to be asymptotically optimal schedule for induction algorithms with polynomial time complexity $\Theta(f(n))$, where $f(n)$ is the expected run-time of the induction algorithm trained with $n$ instances, given that $f$ is no better than $O(n)$ \cite{ps}. This means that, in the worst case scenario, the overall run-time of sequentially running the induction algorithm according to $S_g$ is no worse than running it according to $S_c$. 

The time complexity of a wrapper $CHC$ is linearly dependent on the complexity of the induction algorithm \cite{mapreduce}, and could be expressed as $O(f(n)\cdot e)$, where $e$ is the number of fitness function evaluations. Therefore, for a fixed number of evaluations $e$ on each surrogate level, we could directly extend the proof. The total run-time of running several $CHCs$ sequentially according to schedule $S_g$ for a fixed number of evaluations $e$ on each level is at worst equivalent to running a single $CHC$ according to $S_c$ for the same fixed number of evaluations $e$. The expected number of iterations at different levels is impossible to estimate, however, intuitively, there is no reason any one should be significantly different from the others; and the upper bound is clearly the same.






In the analysis we consider three classifiers of different time complexity orders. Decision Trees $(DT)$, with a linear complexity in number of instances $O(n\cdot k^2)$ \cite{dt-complexity}, k-Nearest Neighbors $(kNN)$, with a quadratic complexity $O(n^2\cdot k)$, and a nonlinear Support Vector Machine $(SVM)$, with a cubic complexity $O(n^3)$ \cite{svm_complexity}.



\begin{figure}[t]
\centerline{\includegraphics[width=0.5\textwidth]{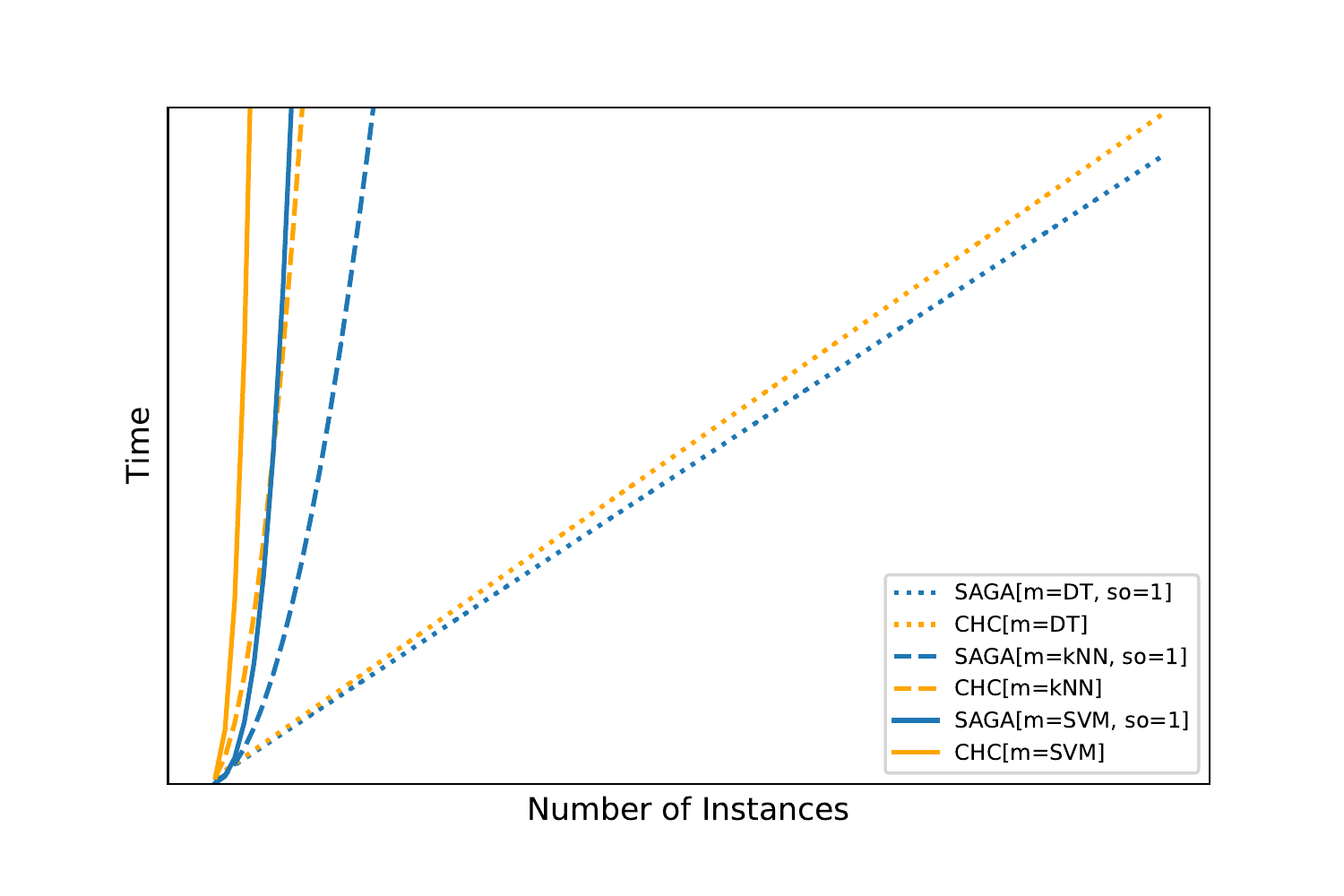}}
\caption{The run-time of $SAGA$ in comparison to a classical wrapper $GA$ for different induction algorithms.}
\label{fig1}
\end{figure}

To visualize this intuition, we plot the expected time required to run all surrogates sequentially according to $S_g$ for a fixed number of evaluations $e$, along with a classical $CHC$ wrapper running according to $S_c$, in Fig.~\ref{fig1}. As observed from the plot, the advantage of using $SAGA$ progressive sampling approach increases as the complexity of the underlying induction algorithm grows. The percentage of time $SAGA$ requires in comparison to the $CHC$ time is: $93.75\%$, $33.20\%$ and $14.28\%$, for the $DT$, $kNN$ and $SVM$ classifiers, respectively. 

\subsection{Evolution Control and False Optimum Prevention} 
Surrogates have been found to often mislead the evolutionary algorithm towards false optima \cite{falseoptimum}. A false optimum is defined as a point which is an optimum of the approximation, but not of the original fitness function \cite{sampling-survey}. This problem is more serious in high dimensional datasets, where constructing a surrogate that offers a perfect global approximation of the original function is inherently harder. Therefore, a strategy (also known as evolution control or model management \cite{sa-survey}) of using the original function to prevent the surrogate from misleading the search is often adopted. The evolution control in $SAGA$ is designed to manage surrogates, and govern the decisions to switch from one level to the next on the basis of reducing the chance of falling to a false optima. 


\subsection{Individuals Migration Strategy}
The multi-level surrogate architecture of $SAGA$ requires a strategy to migrate solutions between different levels of surrogates. Our strategy is designed to always pass the best found individual to the next level surrogate. Once the surrogate stage is over, $SAGA$ migrates the best found individual $(g^\prime)$ to the final population, and uses $(g^\prime)$ as a seed to create the other individuals. 

Additionally, $SAGA$ accept Surrogate Perseverance $(sp)$ hyper-parameter which allows the algorithm to run a number of surrogates of the same level before moving to the next level. Whenever $(sp>1)$, only the best solution based on true fitness out of candidates found by the current level surrogates is migrated to the next level.

\subsection{Population Size Reduction}
$SAGA$ is designed to progressively reduce the population size from one surrogate level to the next. The intuition behind introducing this is based on the hypothesis that as the algorithm progresses from one surrogate to the next, the need to perform exploration gradually decreases. This hypothesis is based on the fact that the algorithm always migrates the best found solution to the next level population. Therefore, a large population that allows more possibilities for exploration should not be as needed for an advanced level surrogate as a near-optimal solution already exists in the initial population.

\begin{algorithm}
    \LinesNumbered
    \SetKwFunction{FeatureSelection}{FeatureSelection}
    \SetKwInOut{KwIn}{Input}
    \SetKwInOut{KwOut}{Output}
    \KwIn{$(m)$ machine learning model.\\$(p)$ number of individuals in population.\\$(k)$ number of generations.\\$(c_p)$ crossover probability.\\}
    \KwOut{$(g^*)$ Best found feature subset.}
    
    \tcp{Initialize a population. 
    }
    $InitPopulation[P(t), p];$
    
    \tcp{Evaluate population using $m$.}
    $EvalPopulation[P(t), m];$ 
    
    $t \leftarrow 1$\;
    \While{$t \leq k$}{
  
        \tcp{Select Parents.}
        $P\prime(t) \leftarrow SelectCHC[P(t)];$ \

        \tcp{Mate parents (preventing incest).}
        $C(t) \leftarrow\ CrossCHC[P\prime(t)];$ 
        
        \tcp{Evaluate offsprings.}
        $EvalPopulation[C(t)];$ 
        
        $t \leftarrow t+1$\;
        
        \tcp{Form a new generation.}
        $P(t) \leftarrow SelectBest[P(t), C(t), j]$;

  \If{$Convergence[P(t)]$} {
            \tcp{Use the best feature subset $g\prime$ for reinitialization.}
            $g\prime \leftarrow SelectBest[P(t), 1]$\;
            $Reinitialize[P(t), g\prime]$
            }
 }
 
    \tcp{Select the best feature subset.}
    $g^* \leftarrow SelectBest[P(t), 1]$\;
    
    \KwRet{$g^*$}
    \caption{CHC Feature Selection Algorithm}
    \label{alg:chc}
    \end{algorithm}

\begin{algorithm}
    \LinesNumbered
    \SetKwFunction{FeatureSelection}{FeatureSelection}
    \SetKwInOut{KwIn}{Input}
    \SetKwInOut{KwOut}{Output}
    \KwIn{$(b)$ number of surrogate levels.\\$(pr)$ population reduction rate.\\$(z)$ generation step size.\\$(fop)$ false optimum prevention.\\$(sp)$ surrogate perseverance.\\$(so)$ surrogate stage only.\\$(D)$ training dataset.\\$(m)$ machine learning model.\\$(p0)$ number of individuals in population.\\$(k)$ number of generations.\\$(c_p)$ crossover probability.\\$(m_p)$ mutation probability.\\}
    \KwOut{$(g^*)$ Best found feature subset.}
    
   \tcp{Sample starting dataset.}
            $s \leftarrow Random[D, \frac{len(D)}{2^b}]$\;
            $p \leftarrow p0$\;
            $switch \leftarrow False$\;
            $persev\_counter \leftarrow sp$\;
    
    \While{True}{
    
        \tcp{Run $z$ generations using $m^*$.} 
        $g\prime \leftarrow GA[m*, p, z, c_p, m_p]$\;
        
        \If{fop AND original fitness is degrading}{
        $switch \leftarrow True$}
        
        \If{current surrogate converged OR switch} {
            $persev\_counter \leftarrow persev\_counter - 1$\;
            \If{$persev\_counter == 0$}{\tcp{Switch to next level.}
            $b \leftarrow b-1$\;
            $persev\_counter \leftarrow sp$\;
            }

            \If{$b == 0$}{
               \tcp{Surrogate stage is over.}
               \If{$so$}{
               \KwRet{$g^\prime$}
               }
               
               $p \leftarrow p0$\;
               
               \tcp{Use $g\prime$ as a seed to initialize population.}
               $InitPopulation[P(t), p, g\prime];$

                break\;
                }
            
            \tcp{Sample partial training set.}
            $s \leftarrow Random[D, \frac{len(D)}{2^b}]$\;
    
            \tcp{Reduce population size.}
            $p \leftarrow p\times pr$\;
        
           \tcp{Initialize next level  population and migrate $g\prime$.}
           
            $InitPopulation[P(t), p];$
               
            $P(t) \leftarrow SelectBest[P(t), g\prime, p]$;
            $switch \leftarrow False$\;
        }
        
    }
    \tcp{Run $k$ generation using $m$.} 
        $g^* \leftarrow GA[m, D, p, k, c_p, m_p]$\;
            
    \KwRet{$g^*$}
    \caption{SAGA Feature Selection Algorithm}
    \label{alg:saga}
\end{algorithm}

\section{Experiments and Results}
In this section, we explain the experimental design used to evaluate $SAGA$, and present the results of this evaluation. In all experiments, we analyze two conflicting objectives, namely the computation time and accuracy achieved. In other words, we look at how long it takes to find feature subsets, and how good they are. In all experiments, we report the results of t-test to determine significance difference on a number of datasets.  


The initial experiment focuses on the exploration capabilities of surrogates, as compared to those of the original function. To this end, we first run $SAGA$ using only the surrogates to lead the evolutionary search. We then analyze the time it takes for classical $CHC$ to match the obtained accuracy. Finally, we compare those results against running, all the way to convergence, both the $CHC$ and the full version of \SAGA.




In the second experiment, we demonstrate the benefits of the false optimum prevention evolution control of \SAGA.
The final experiments illustrate the sensitivity of the algorithm to the selection of two main hyper-parameters, namely the population reduction rate and surrogate perseverance.

\subsection{Experimental Setup}
A total of 14 datasets from the UCI Machine Learning Repository\footnote{\url{http://archive.ics.uci.edu/ml}} are used in our experiments. We have selected high dimensional datasets of thousands of instances, and hundreds of features. Information about selected datasets is presented in Table~\ref{tab1}. A repository\footnote{\url{https://github.com/Ghaith81/SAGA}} of the framework implementation is available publicly to ensure reproducibility of results.

\begin{table}
\centering
\caption{information of UCI datasets used in experiments.}\label{tab1}
\begin{tabular}{|l|l|l|l|}
\hline
dataset &  No. of Instances & No. of Features &  No. of Classes\\
\hline
dermatology & 366 & 34 & 6\\
\hline
german & 1\,000 & 24 & 2\\
\hline
semeion & 1\,592 & 265 & 2\\
\hline
car & 1\,728 & 6 & 4\\
\hline
abalone & 4\,177 & 8 & 28\\
\hline
qsar & 8\,992 & 1\,024 & 2\\
\hline
epileptic & 11\,500 & 178 & 2\\
\hline
adult & 32\,561 & 14 & 2\\
\hline
bank-full & 45\,211 & 16 & 2\\
\hline
connect-4 & 67\,556 & 42 & 3\\
\hline
dota 2 & 92\,650 & 116 & 2\\
\hline
diabetes & 101\,766 & 49 & 3\\
\hline
census-income & 199\,523 & 41 & 2\\
\hline
covtype & 581\,012 & 54 & 7\\
\hline
\end{tabular}
\end{table}

We have adopted a unified prepossessing approach of all datasets. Encoding of categorical feature is done using \mbox{LabelEncoder} from \texttt{scikit-learn}. Missing values are imputed using the median for continuous features and the mode for categorical ones. All datasets are shuffled, then a fixed split of (60\%) training, (20\%) validation, and (20\%) testing is used across all experiments.

A Decision Tree classifier is used as an induction algorithm throughout this paper, for a conservative choice. The fitness function of $GA$ is always maximizing the accuracy score on the validation split. 
A fixed population size of 40 individuals is used for $CHC$, and the same number of individuals is used as the starting population size $p_0$ of $SAGA$. The convergence criteria is based on a fixed number of generations (set to 10) without the fitness of the best individual in the population improving.

We will use the following terminology to refer to the algorithms discussed in the subsequent sections:   
\begin{itemize}
  \item\textbf{Baseline:} The Decision Tree classifier that is directly trained using all available features (no feature selection).
  \item\textbf{CHC[p=40]:} The Decision Tree classifier that is trained using feature subset selected by $CHC$ feature selection algorithm. The settings in brackets showcase the default hyper-parameter.
  \item\textbf{SAGA[b=4, pr=0.5, z=10, sp=1, so=0, p0=40]:} The Decision Tree classifier that is trained using feature subset selected by $SAGA$ (Algorithm~\ref{alg:saga}). The settings in brackets showcase the default hyper-parameters.
\end{itemize}

\subsection{Experiment I - Performance of SAGA}
\label{exp1}
We carry out the first experiment in two folds. The first part is designed to evaluate the exploration capabilities in terms of speed and fitness of surrogates, as compared to those of the original function. The time aspect is validated by running SAGA[so=1], a version that is limited to the surrogate stage to convergence, and recording the time required by $CHC$ to match the found solution in terms of fitness. We allow $CHC$ to continue running to convergence to compare the quality of solutions found by the surrogate-based method in reference to a classical approach. 
The second fold of the experiment compares the full version of $SAGA$ against $CHC$ by running both algorithms until convergence.

Outcomes of the first part of the experiment are presented in Table~\ref{tab2}.
Two prominent conclusions, in terms of computation time, can be drawn from the results shown in first two columns (Time). First, for large datasets, SAGA[so=1] is always significantly (p-value$=0.001$) faster than $CHC$. In fact, for the 9 largest datasets in our experiments, $CHC$ was at least twice slower, requiring on average $3.15$ times more computations than SAGA[so=1] to arrive at solutions of comparable fitness. Of course, such computational efficiency comes with a drawback in terms of overall accuracy that is possible to achieve.

On the other hand, the surrogate-based approach SAGA[so=1] is statistically significantly slower than a classical $GA$ for relatively small datasets (less than $5\, 000$ instances in our experiments). This observation is easily explained by the overhead of using $SAGA$ algorithm, which involves constructing and switching between multiple levels of surrogates. Such an overhead of the algorithm turns out to be costly in terms of time in the case of small datasets. At the same time, the difference is on the order of one or two seconds on a modern laptop, so unless the feature selection is performed a massive number of times, or using a very limited hardware capacity, the difference is unlikely to matter.

As expected, in terms of solution quality, as demonstrated in the last two columns of Table~\ref{tab2} (Accuracy), feature subsets found by SAGA[so=1] are statistically significantly (p-value$=0.0162$) worse than the classical $GA$. The latter has always converged to a solution resulting in higher classification accuracy. On average, $CHC$ identified solutions that are $0.86\%$ better than those found by SAGA[so=1]. This demonstrates the downside of using only the surrogates to lead the optimization search; the search converges much faster, however, it selects sub-optimal solutions.

The observations of this experiment explain the fundamental concept of our algorithm. A surrogate-based approach is beneficial during the exploratory phase, as it leads the feature selection search to near-optimal solutions faster than the classical wrapper approach. However, the approximation ends up having too low fidelity, which is a critical impediment in terms of exploitation. Therefore, our complete algorithm, SAGA[so=0], switches to the original function at the very end, for full exploitation once the surrogates has reached the limit.  

The second fold of the experiment compares the performance of the full version of $SAGA$, which utilizes the surrogates for exploration and the original function for exploitation, against $CHC$ feature selection, as well as the baseline Decision Tree Classifier (without performing feature selection). In this setting, both $SAGA$, and $CHC$ run until convergence. The goal of this experiment is to analyze the computation time required by each feature selection approach to achieve the maximum solution quality possible.

\begin{figure*}[htbp]
  \begin{subfigure}[b]{1\columnwidth}
    \includegraphics[width=\linewidth]{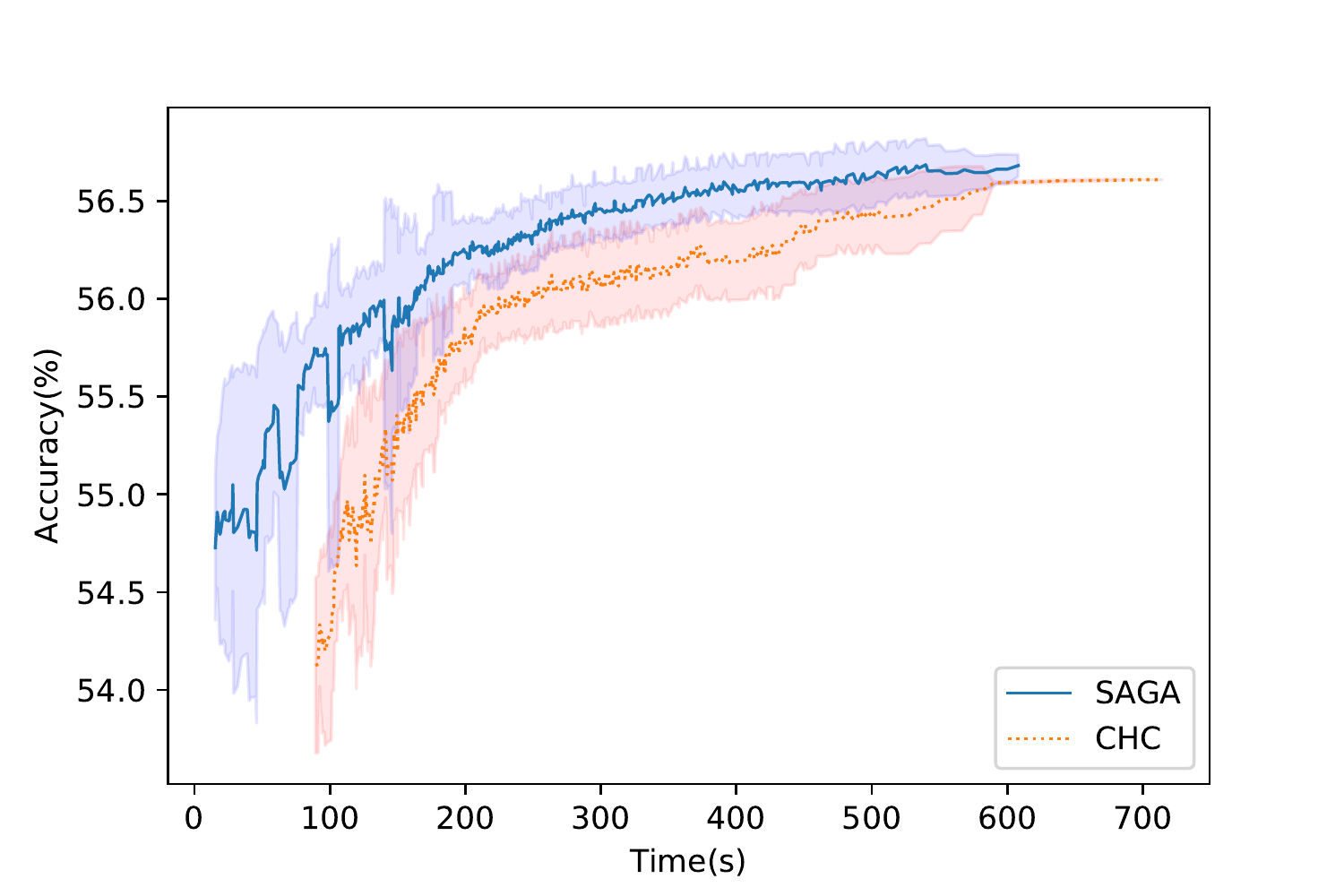}
    \caption{dota2Train}
  \end{subfigure}
  \hfill 
  \begin{subfigure}[b]{1\columnwidth}
    \includegraphics[width=\linewidth]{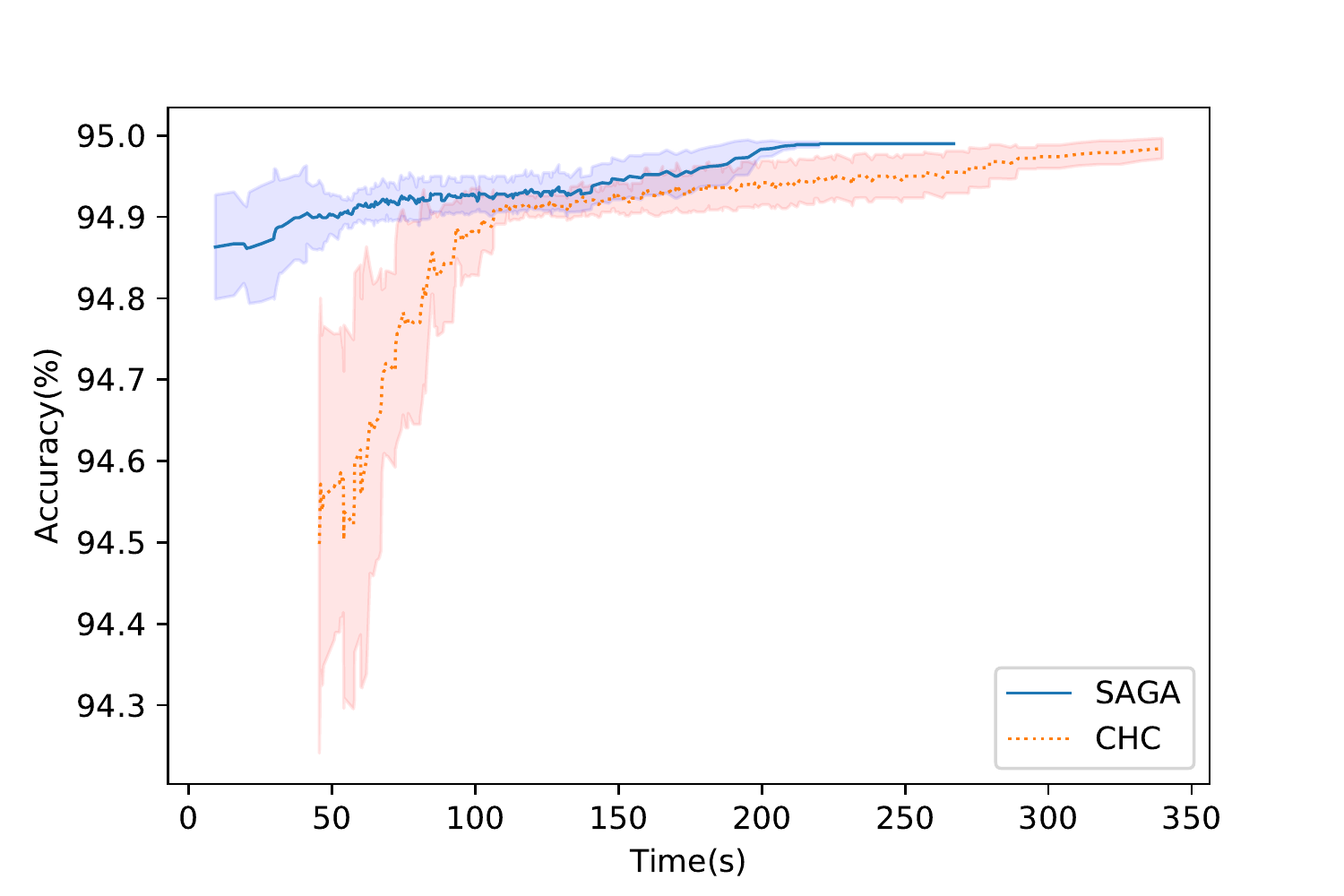}
    \caption{census-income}
    \label{fig:2}
  \end{subfigure}
  \caption{A visualization of the 10 runs of both $SAGA$ and $CHC$ reported in Table~\ref{tab3}.}
\end{figure*}

Results of the second fold of this experiment are presented in Table~\ref{tab3}. We may observe from computation time results shown in the first two columns (Time) that $SAGA$ was faster than $CHC$ in 6 out of 9 large datasets. $SAGA$ was significantly faster than $CHC$ -- on average it only required $84.68\%$ of $CHC$ time (p-value$=0.001$). The last three columns of Table~\ref{tab3} (Accuracy) shows that $SAGA$ outperformed $CHC$ in 9 out of 14 datasets. $SAGA$ found feature subset solutions that are significantly higher in terms of accuracy, on average by $0.24\%$ (p-value$=0.0496$) in comparison to solutions found by $CHC$. We also report the improvement in accuracy realized by $SAGA$ in reference to $Baseline$. The feature subsets found by $SAGA$ averaged $3.76\%$ (p-value$=0.0059$) higher accuracy in comparison to not performing feature selecting. $CHC$ averaged $3.52\%$ (p-value$=0.0053$) accuracy improvement. We visualize the progress over time of the feature selection search for two datasets ({\tt dota2Train} and {\tt census-income}), using both $SAGA$ and $CHC$, in Figure~\ref{fig:2}. The progress of each algorithm is shown by applying a smoothing moving average with a window of 10 to the highest accuracy found. The shaded range in the figure is the standard deviation. Evidently, $SAGA$ starts producing high quality solutions much earlier than $CHC$, and ultimately converges to better solutions. 

These experiments show that $SAGA$ does not trade off accuracy to reduce computation time. In fact, $SAGA$ is shown to outperform a classical wrapper $GA$ on both conflicting objectives of time and accuracy. The surrogate-assisted algorithm reduces the computation of the wrapper $GA$, while still improving the accuracy. 


Both parts of this experiment together demonstrate that a surrogate-based evolutionary approach for wrapper feature selection is more efficient computationally than a classical evolutionary approach during exploration. We expressly show that surrogates have a limited exploitation capability in comparison to an evolutionary approach utilizing the original function. Our framework is designed, accordingly, to carry exploration using surrogates, and only switch to the original function at a later stage of the optimization search. Our surrogate-assisted algorithm $SAGA$ is significantly faster, at least for large datasets (over 5\,000 instances), and generally converges to feature subset solutions of significantly higher accuracy.

\begin{table*}[htbp]
\caption{an approach using surrogates only for Feature Selection in comparison to CHC, best algorithm is in bold.}
\begin{center}
\begin{tabular}{|c|c|c|c|c|}
\hline
\textbf{Dataset}&\multicolumn{2}{|c|}{\textbf{Time} (seconds)} &\multicolumn{2}{|c|}{\textbf{Accuracy}}\\
\cline{2-5} 
\textbf{} & \textit{\textbf{SAGA$[so=1]$}}& \textbf{\textit{\textbf{CHC}$^\mathrm{a}$}}& \textbf{\textit{\textbf{SAGA$[so=1]$}}}& \textbf{\textit{\textbf{CHC}}} \\

\hline
dermatology & 1.11$ \pm 0.45$&{\bfseries 0.10}$ \pm 0.05$&{97.12\%}$ \pm0.78$&{\bfseries 99.18\%}$ \pm0.96$\\
\hline
german & 1.16$ \pm 0.45$&{\bfseries 0.22}$  \pm 0.23$&{72.10\%}$ \pm1.60$&{\bfseries 74.85\%}$ \pm0.47$\\
\hline
semeion & 1.94$ \pm 1.08$&{\bfseries 1.12}$  \pm 1.06$&{94.37\%}$ \pm1.00$&{\bfseries 96.98\%}$ \pm0.56$\\
\hline
car & 1.27$ \pm 0.20$&{\bfseries 0.09}$  \pm 0.01$&{94.80\%}$ \pm0.86$&{\bfseries 95.38\%}$ \pm0.00$\\
\hline
abalone & 2.05$ \pm 0.99$&{\bfseries 1.27}$  \pm 1.00$&{24.35\%}$ \pm3.08$&{\bfseries 26.81\%}$ \pm1.32$\\
\hline
qsar & {\bfseries31.08}$ \pm 15.02$&{304.25}$  \pm 797.58$&{91.74\%}$ \pm0.37$&{\bfseries 92.64\%}$ \pm0.21$\\
\hline
epileptic & {\bfseries52.17}$ \pm 27.93$&{164.47}$  \pm 169.09$&{94.38\%}$ \pm0.48$&{\bfseries 95.64\%}$ \pm0.29$\\
\hline
adult & {\bfseries3.32}$ \pm 1.60$&{7.81}$  \pm 1.76$&{85.94\%}$ \pm0.14$&{\bfseries 86.01\%}$ \pm0.02$\\
\hline
bank-full & {\bfseries3.10}$ \pm 1.62$&{9.48}$  \pm 7.13$&{89.92\%}$ \pm0.02$&{\bfseries 89.95\%}$ \pm0.02$\\
\hline
connect-4 & {\bfseries31.42}$ \pm 7.91$&{87.69}$  \pm 56.34$&{76.86\%}$ \pm0.75$&{\bfseries 77.88\%}$ \pm0.08$\\
\hline
dota2Train & {\bfseries68.60}$ \pm 22.30$&{426.11}$  \pm 567.04$&{55.92\%}$ \pm0.25$&{\bfseries 56.23\%}$ \pm0.27$\\
\hline
diabetic & {\bfseries24.16}$ \pm 7.62$&{106.40}$  \pm 58.21$&{57.49\%}$ \pm0.34$&{\bfseries 57.77\%}$ \pm0.17$\\
\hline
census-income & {\bfseries25.87}$ \pm 13.99$&{103.77}$  \pm 34.13$&{94.90\%}$ \pm0.02$&{\bfseries 94.93\%}$ \pm0.03$\\
\hline
covtype & {\bfseries606.97}$ \pm 158.66$&{1264.76}$  \pm 705.46$&{93.69\%}$ \pm0.10$&{\bfseries 93.88\%}$ \pm0.03$\\
\hline

\multicolumn{4}{l}{$^{\mathrm{a}}$The time required by $CHC$ to match solutions found by $SAGA[so=1]$.}
\end{tabular}
\label{tab2}
\end{center}
\end{table*}

\begin{table*}[htbp]
\caption{SAGA Feature Selection Performance in comparison to CHC.}
\begin{center}
\begin{tabular}{|c|c|c|c|c|c|}
\hline
\textbf{Dataset}&\multicolumn{2}{|c|}{\textbf{Time} (seconds)} &\multicolumn{3}{|c|}{\textbf{Accuracy}}\\
\cline{2-6} 
\textbf{} & \textit{\textbf{SAGA}}& \textbf{\textit{\textbf{CHC}}}& \textbf{\textit{\textbf{SAGA}}}& \textbf{\textit{\textbf{CHC}}}&\textbf{\textit{\textbf{Baseline}}} \\

\hline
dermatology & 1.90$ \pm 0.27$&{\bfseries 0.45}$ \pm 0.22$&{\bfseries 99.32\%}$ \pm0.72$&{99.18\%}$ \pm0.96$&95.38\%\\
\hline
german & 2.73$ \pm 0.35$&{\bfseries 1.67}$  \pm 0.45$&{\bfseries 76.40\%}$ \pm1.07$&{74.85\%}$ \pm0.47$&73.00\%\\
\hline
semeion & 11.99$ \pm 3.51$&{\bfseries 9.02}$  \pm 2.39$&{\bfseries 97.55\%}$ \pm0.49$&{96.98\%}$ \pm0.56$&94.03\%\\
\hline
car & 2.16$ \pm 0.12$&{\bfseries 0.70}$  \pm 0.02$&{95.38\%}$ \pm0.00$&{95.38\%}$ \pm0.00$&95.38\%\\
\hline
abalone & 4.80$ \pm 0.47$&{\bfseries 3.78}$  \pm 1.12$&{\bfseries 27.43\%}$ \pm0.00$&{ 26.81\%}$ \pm1.32$&20.60\%\\
\hline
qsar & {\bfseries262.09}$ \pm 15.02$&{426.43}$  \pm 133.99$&{\bfseries 93.14\%}$ \pm0.44$&{92.64\%}$ \pm0.21$&89.94\%\\
\hline
epileptic & {\bfseries899.79}$ \pm 27.93$&{1344.53}$  \pm 613.33$&{95.46\%}$ \pm0.20$&{\bfseries 95.64\%}$ \pm0.29$&93.09\%\\
\hline
adult & {17.62}$ \pm 3.57$&{\bfseries17.60}$  \pm 5.54$&{86.01\%}$ \pm0.03$&{86.01\%}$ \pm0.02$&80.22\%\\
\hline
bank-full & {\bfseries15.55}$ \pm 2.32$&{18.18}$  \pm 3.43$&{89.95\%}$ \pm0.03$&{89.95\%}$ \pm0.02$&87.22\%\\
\hline
connect-4 & {\bfseries152.04}$ \pm 36.18$&{200.88}$  \pm 29.91$&{77.87\%}$ \pm0.09$&{\bfseries 77.88\%}$ \pm0.08$&72.24\%\\
\hline
dota2Train & {423.98}$ \pm 118.59$&{\bfseries418.97}$  \pm 136.26$&{\bfseries 56.54\%}$ \pm0.15$&{56.23\%}$ \pm0.27$&51.77\%\\
\hline
diabetic & {\bfseries122.53}$ \pm 53.51$&{146.22}$  \pm 41.11$&{\bfseries 57.80\%}$ \pm0.13$&{57.77\%}$ \pm0.17$&48.83\%\\
\hline
census-income & {\bfseries160.62}$ \pm 56.86$&{227.08}$  \pm 68.87$&{\bfseries 94.94\%}$ \pm0.03$&{94.93\%}$ \pm0.03$&92.85\%\\
\hline
covtype & {5385.19}$ \pm 1298.26$&{\bfseries4614.19}$  \pm 874.28$&{\bfseries 93.91\%}$ \pm0.06$&{93.88\%}$ \pm0.03$&93.00\%\\
\hline

\multicolumn{4}{l}{}
\end{tabular}
\label{tab3}
\end{center}
\end{table*}

\subsection{Experiment II - False Optimum Prevention}
In this experiment, we evaluate the effectiveness of our evolution control approach of preventing false optimum, which has been recognized as the main problem of low fidelity approximations \cite{falseoptimum}. We compare version SAGA[so=1, fop=1] following our evolution control against a version of SAGA[so=1, fop=0] that essentially employs no evolution control. This version runs each surrogate all the way to convergence before switching to the next, disregarding the possibility of false optima. From this point forward, we run all experiments of $SAGA$ with [so=1], as we are interested in analyzing the impact during the surrogate-assisted stage only.

Results of this experiment are shown in Table~\ref{tab4}. A notable conclusion in terms of computation time, from the first two columns (Time), is that SAGA[so=1, fop=1] is faster than version SAGA[so=1, fop=0] on 8 out of 14 datasets. Although the false optimum prevention requires individual evaluations using the expensive original function, the time saved this way usually outweighs the cost. Our strategy will often abandon a surrogate early, whenever it fails to improve the original fitness for $z$ generations. Our evolution control is shown to reduce the computation time in 7 out of the largest 9 datasets (larger than 5\, 000 instances). For large datasets SAGA[so=1, fop=1] required significantly shorter time compared to SAGA[so=1, fop=0], on average $84.17\%$ (p-value$=0.0476$). Similarly to what we found in the previous experiment for the case of small datasets, the overhead outplays the benefit. The difference is unlikely to matter as is on the order of a second. 

In terms of accuracy, the false optimum prevention evolution control achieved better accuracy in 10 out of 14 datasets. SAGA[so=1,fop=1] produced solutions of a marginally higher (p-value$=0.1109$) average accuracy by $0.49\%$ compared to SAGA[so=1,fop=0]. The time and accuracy results show that our evolution control strategy effectively reduced the negative effect of surrogates misdirecting the search towards a false optimum, as observed by higher average accuracy, and significantly shorter time. 

\begin{table*}[htbp]
\caption{evaluation of $SAGA$ evolution control effect on time and accuracy.}
\begin{center}
\begin{tabular}{|c|c|c|c|c|}
\hline
\textbf{Dataset}&\multicolumn{2}{|c|}{\textbf{Time} (seconds)} &\multicolumn{2}{|c|}{\textbf{Accuracy}}\\
\cline{2-5} 
\textbf{} & \textit{\textbf{SAGA$[so=1, fop=1]$}}& \textbf{\textit{\textbf{SAGA$[so=1, fop=0]$}}}& \textbf{\textit{\textbf{SAGA$[so=1, fop=1]$}}}& \textbf{\textit{\textbf{SAGA$[so=1, fop=0]$}}} \\

\hline
dermatology & 1.50$ \pm 0.21$&{\bfseries 1.22}$  \pm 0.20$&{\bfseries 97.12\%}$ \pm1.20$&{95.89\%}$ \pm2.42$\\
\hline
german & {\bfseries 1.50}$ \pm 0.18$&1.82$  \pm 0.46$&{\bfseries 72.35\%}$ \pm1.06$&{70.35\%}$ \pm1.68$\\
\hline
semeion & 3.09$ \pm 0.49$&{\bfseries 2.91}$  \pm 0.53$&{\bfseries 93.77\%}$ \pm0.82$&{92.58\%}$ \pm1.23$\\
\hline
car & 1.66$ \pm 0.13$&{\bfseries 1.06}$  \pm 0.10$&{\bfseries 94.68\%}$ \pm0.90$&{94.51\%}$ \pm0.92$\\
\hline
abalone & 3.15$ \pm 0.58$&{\bfseries 2.36}$\pm 0.43$&{\bfseries 24.67\%}$ \pm3.28$&{22.75\%}$ \pm3.49$\\
\hline
qsar & 57.15$ \pm 27.46$&{\bfseries 48.24}$  \pm 17.85$&{91.35\%}$ \pm0.06$&{\bfseries 91.49\%}$ \pm0.73$\\
\hline
epileptic & {\bfseries 85.65}$ \pm 27.07$&118.59 $  \pm 22.56$&{\bfseries 94.18\%}$ \pm0.37$&{93.3\%}$ \pm0.50$\\
\hline
adult & {\bfseries 5.89}$ \pm 1.13$&5.97 $  \pm 1.20$&{\bfseries 85.97\%}$ \pm0.02$&{85.96\%}$ \pm0.03$\\
\hline
bank-full & 5.35$ \pm 0.75$&{\bfseries 5.06}$  \pm 0.48$&{\bfseries 89.91\%}$ \pm0.02$&{89.90\%}$ \pm0.02$\\
\hline
connect-4 & {\bfseries 30.37}$ \pm 7.63$&35.9$  \pm 6.55$&{76.45\%}$ \pm0.83$&{\bfseries 76.73\%}$ \pm1.10$\\
\hline
dota2Train &{\bfseries 88.10}$ \pm 30.50$&118.14$  \pm 22.55$&{\bfseries 55.59\%}$ \pm0.51$&{55.55\%}$ \pm0.23$\\
\hline
diabetic &{\bfseries 32.02}$ \pm 5.97$&38.04$  \pm 4.84$&{57.36\%}$ \pm0.35$&{\bfseries 57.58\%}$ \pm0.25$\\
\hline
census-income &{\bfseries 39.43}$ \pm 3.83$&47.92$  \pm 6.80$&{94.90\%}$ \pm0.02$&{94.90\%}$ \pm0.01$\\
\hline
covtype &{\bfseries 790.04}$ \pm 132.41$&992.29$  \pm 153.25$&{\bfseries 93.76\%}$ \pm0.11$&{93.68\%}$ \pm0.10$\\
\hline

\multicolumn{4}{l}{}
\end{tabular}
\label{tab4}
\end{center}
\end{table*}

\subsection{Experiment III - Population Reduction Rate}
\label{prr}
This experiment is designed to analyze the impact of the population reduction rate hyper-parameter on computation time and fitness of solutions. We compare the results of two versions of $SAGA$ here. The first, default, SAGA[so=1, pr=0.5] uses a population reduction rate of $(pr=0.5)$; the other is version of SAGA[so=1, pr=1], i.e., with no population reduction. Naturally, one expects the version with population reduction to run faster. However, the tendency for premature convergence is shown to be inversely proportional to the population size in a classical $GA$ \cite{premature}. Similarly, $SAGA$ with population reduction has a higher risk of premature convergence. This could happen in our framework if an early surrogate introduces a sub-optimal solution that the following levels fail to improve upon due to gradually decreasing exploration possibilities caused by the smaller population. This experiment allows us to analyze the trade off between the chance of premature convergence and computation time.  

The version SAGA[so=1, pr=0.5] starts with a population of 40 individuals using a  surrogate built from $6.25\%$ of available data. The schedule of population size for SAGA[so=1, pr=0.5] is: $\{40, 20, 10, 5\}$, while SAGA[so=1, pr=1] runs a population size schedule of $\{40, 40, 40, 40\}$. Both versions follow a sampling schedule of $S_g = \{6.25\%, 12.50\%, 25.00\%, 50.00\% \}$.

The experiment results of Table~\ref{tab5} shows that the computation time of SAGA[so=1, pr=0.5] is significantly shorter than SAGA[so=1, pr=1] for all tested datasets. On average, SAGA[so=1, pr=0.5] required only $41.73\%$ of SAGA[so=1, pr=1] convergence time. However, in terms of quality of solutions, SAGA[so=1, pr=1] found feature subsets that are of marginally higher (p-value$=0.1597$) accuracy (by $0.21\%$ on average). As expected, the use of the population reduction rate lowers the algorithm computation time at the expense of an increased chance of premature convergence. The difference in time, however, is much higher than the difference in accuracy. Therefore, we recommend choosing a larger initial population $(p_0)$, along with a population reduction rate of $pr=0.5$ for datasets of a large number of features, or in situations of which computation cost is a secondary concern, over $pr=1$. 

\begin{table*}[htbp]
\caption{population reduction rate effect on $SAGA$ time and accuracy performance.}
\begin{center}
\begin{tabular}{|c|c|c|c|c|}
\hline
\textbf{Dataset}&\multicolumn{2}{|c|}{\textbf{Time} (seconds)} &\multicolumn{2}{|c|}{\textbf{Accuracy}}\\
\cline{2-5} 
\textbf{} & \textit{\textbf{SAGA$[so=1, pr=0.5]$}}& \textbf{\textit{\textbf{SAGA$[so=1, pr=1]$}}}& \textbf{\textit{\textbf{SAGA$[so=1], pr=0.5$}}}& \textbf{\textit{\textbf{SAGA$[so=1, pr=1]$}}} \\

\hline
dermatology & {\bfseries 0.98}$ \pm 0.46$&{1.60}$  \pm 0.77$&{\bfseries 97.16\%}$ \pm1.37$&{97.15\%}$ \pm   0.18$\\
\hline
german & {\bfseries 1.21}$ \pm 0.51$&{2.11}$  \pm 0.64$&{\bfseries 73.05\%}$ \pm1.07$&{72.70\%}$ \pm  2.04$\\
\hline
semeion & {\bfseries 1.88}$ \pm 0.93$&{4.40}$  \pm 2.52$&{\bfseries 94.37\%}$ \pm 0.68$&{94.09\%}$ \pm  1.35$\\
\hline
car & {\bfseries 1.27}$ \pm 0.21$&{1.80}$  \pm 0.47$&{\bfseries 94.94\%}$ \pm 0.79$&{94.91\%}$ \pm  0.81$\\
\hline
abalone & {\bfseries2.18}$ \pm 0.58$&{4.68}$  \pm 1.78$&{25.39\%}$ \pm2.44$&{\bfseries 26.57\%}$ \pm1.39$\\
\hline
qsar & {\bfseries 23.8}$ \pm 10.87$&{304.25}$  \pm 797.58$&{91.38\%}$ \pm0.38$&{\bfseries 91.95\%}$ \pm0.74$\\
\hline
epileptic & {\bfseries 42.65}$ \pm 19.06$&{111.67}$  \pm 118.35$&{94.20\%}$ \pm0.39$&{\bfseries 94.22\%}$ \pm0.32$\\
\hline
adult & {\bfseries 3.52}$ \pm 1.71$&{7.37}$  \pm 4.72$&{85.93\%}$ \pm0.11$&{\bfseries 85.98\%}$ \pm0.02$\\
\hline
bank-full & {\bfseries 1.93}$ \pm 1.15$&{8.23}$  \pm 5.23$&{89.91\%}$ \pm0.01$&{\bfseries 89.92\%}$ \pm0.02$\\
\hline
connect-4 & {\bfseries 33.18}$ \pm 7.55$&{103.43}$  \pm 22.02$&{76.01\%}$ \pm1.21$&{\bfseries 77.63\%}$ \pm0.64$\\
\hline
dota2Train & {\bfseries 79.21}$ \pm 51.31$&{175.4}$  \pm 76.57$&{55.77\%}$ \pm0.33$&{\bfseries56.06\%}$ \pm0.25$\\
\hline
diabetic & {\bfseries 24.57}$ \pm 6.45$&{69.99}$  \pm 28.2$&{57.54\%}$ \pm0.27$&{\bfseries 57.62\%}$ \pm0.08$\\
\hline
census-income & {\bfseries 22.43}$ \pm 13.11$&{54.79}$  \pm 40.08$&{94.90\%}$ \pm0.02$&{\bfseries 94.91\%}$ \pm0.02$\\
\hline
covtype & {\bfseries 516.09}$ \pm 323.3$&{2122.09}$  \pm 799.71$&{93.73\%}$ \pm0.07$&{\bfseries 93.78\%}$ \pm0.08$\\
\hline

\multicolumn{4}{l}{}
\end{tabular}
\label{tab5}
\end{center}
\end{table*}

 \subsection{Experiment IV - Individuals Migration Strategy}
 \label{sp}
 In this experiment, we analyze the effect of re-running the same surrogate level more than once, using the hyper-parameter ``surrogate perseverance'' ($sp$). We compare the time and accuracy results of a version using the default setting $(sp=1)$ against $(sp=2)$. 
 
 As observed in the previous experiment, SAGA[so=1, sp=1] follows a population size schedule of $\{40, 20, 10, 5\}$ (for a starting population $p0=40$). In case of SAGA[so=1, sp=2], we start with 40, followed by another 40, at the same surrogate level, i.e., both times using $6.25\%$ data instances. This is then followed by a population size of 20, then another 20, and so on. Upon completion of each level, we migrate the best individual of the two runs according to the original function to the next level population. Again, one expects the $(sp=1)$ version to be faster. However, a version running $(sp=2)$ reduces the chance of premature convergence, choosing the best of two feature subsets according to their original function fitness on each surrogate level.    
 
 The results in Table~\ref{tab6} show that the computation time of the version running $(sp=1)$ is, as expected, significantly shorter for all tested datasets. On average, SAGA[so=1] only required $56.89\%$ of the time SAGA[so=1, sp=2] took to converge. In terms of solutions fitness, SAGA[so=1, sp=2] arrived at solutions $0.47\%$ higher (p-value$=0.2471$) in accuracy. We believe the higher accuracy indicates that increasing surrogate perseverance reduced the chance of premature convergence at the expense of increased computation time. This experiment, along with Experiment~\ref{prr} demonstrates possibilities to revise $SAGA$ hyper-parameters to more efficiently trade off the conflicting objectives of time and accuracy. In applications when computation time is not a priority, a larger initial population $(p_0)$, and surrogate perseverance $sp>1$ could be considered, to increase the possibility of finding feature subset solutions of higher accuracy.

 \begin{table*}[htbp]
\caption{Surrogate perseverance effect on $SAGA$ time and accuracy performance.}
\begin{center}
\begin{tabular}{|c|c|c|c|c|}
\hline
\textbf{Dataset}&\multicolumn{2}{|c|}{\textbf{Time} (seconds)} &\multicolumn{2}{|c|}{\textbf{Accuracy}}\\
\cline{2-5} 
\textbf{} & \textit{\textbf{SAGA$[so=1, sp=1]$}}& \textbf{\textit{\textbf{SAGA$[so=1, sp=2]$}}}& \textbf{\textit{\textbf{SAGA$[so=1, sp=1]$}}}& \textbf{\textit{\textbf{SAGA$[so=1, sp=2]$}}} \\

\hline
dermatology & {\bfseries 1.30}$ \pm 0.13$&{2.21}$  \pm 0.22$&{ 97.12\%}$ \pm1.88$&{\bfseries 97.40\%}$ \pm   1.01$\\
\hline
german & {\bfseries 1.50}$ \pm 0.38$&{2.57}$  \pm 0.36$&{71.85\%}$ \pm1.55$&{\bfseries72.50\%}$ \pm  0.79$\\
\hline
semeion & {\bfseries 3.57}$ \pm 0.67$&{5.77}$  \pm 0.77$&{94.59\%}$ \pm 0.55$&{\bfseries 94.87\%}$ \pm 1.07$\\
\hline
car & {\bfseries 1.46}$ \pm 0.13$&{2.69}$  \pm 0.26$&{94.51\%}$ \pm 0.92$&{\bfseries 95.03\%}$ \pm 0.73$\\
\hline
abalone & {\bfseries2.89}$ \pm 0.42$&{4.28}$  \pm 0.60$&{22.52\%}$ \pm1.79$&{\bfseries 26.83\%}$ \pm1.26$\\
\hline
qsar & {\bfseries 46.69}$ \pm 17.90$&{92.22}$  \pm 19.09$&{\bfseries 91.71\%}$ \pm0.68$&{91.65\%}$ \pm0.32$\\
\hline
epileptic & {\bfseries 84.11}$ \pm 11.69$&{144.47}$  \pm 22.87$&{94.19\%}$ \pm0.25$&{\bfseries 94.31\%}$ \pm0.20$\\
\hline
adult & {\bfseries 5.88}$ \pm 0.86$&{10.01}$  \pm 1.17$&{85.97\%}$ \pm0.03$&{\bfseries 85.98\%}$ \pm0.03$\\
\hline
bank-full & {\bfseries 5.27}$ \pm 0.49$&{10.91}$  \pm 1.11$&{89.91\%}$ \pm0.02$&{\bfseries 89.92\%}$ \pm0.02$\\
\hline
connect-4 & {\bfseries 32.42}$ \pm 8.04$&{49.71}$  \pm 7.06$&{76.28\%}$ \pm0.89$&{\bfseries 76.47\%}$ \pm0.88$\\
\hline
dota2Train & {\bfseries 67.24}$ \pm 15.87$&{160.48}$  \pm 39.66$&{55.76\%}$ \pm0.16$&{\bfseries55.90\%}$ \pm0.23$\\
\hline
diabetic & {\bfseries 33.27}$ \pm 3.68$&{57.96}$  \pm 7.99$&{\bfseries 57.58\%}$ \pm0.25$&{57.56\%}$ \pm0.23$\\
\hline
census-income & {\bfseries 46.09}$ \pm 7.30$&{79.73}$  \pm 11.97$&{94.91\%}$ \pm0.02$&{94.91\%}$ \pm0.02$\\
\hline
covtype & {\bfseries 731.43}$ \pm 156.17$&{1269.25}$  \pm 184.09$&{93.62\%}$ \pm0.17$&{\bfseries 93.79\%}$ \pm0.09$\\
\hline

\multicolumn{4}{l}{}
\end{tabular}
\label{tab6}
\end{center}
\end{table*}


\section{Conclusions}
In this paper, we propose a general framework to perform surrogate-assisted wrapper feature selection, and $SAGA$, an algorithm using $GA$ designed based on this framework. We divide the evolutionary search into the distinctive phases of exploration and exploitation. $SAGA$ utilizes surrogates constructed according to a progressively larger sample size during exploration, and only switches to the original function at the final stage of exploitation. We show experimentally that $SAGA$ can arrive at near-optimal feature subset solutions three times faster than a classical wrapper $GA$. The comparison between our surrogate-assisted approach and a classical $GA$ shows that $SAGA$ converges significantly faster to feature subsets solutions of significantly higher accuracy.

We demonstrate that the evolution control of $SAGA$ reduces the chance of surrogates misleading the search towards false optima by converging to solutions of higher average fitness in a significantly shorter time. Finally, we investigate fine-tuning key hyper-parameters of $SAGA$ to prioritize one of the two conflicting objectives of accuracy and computational cost.

\section{Limitations and Future Work}
 The strategy of gradually decreasing population size in our framework is designed to favor exploration at early stages. However, it is precisely then that the surrogate evaluation is less reliable, with a higher risk of "false good" solutions. A threshold of what qualifies as a "good" solution could be incorporated into our framework to disregard "false good" solutions, and to prevent an early unreliable surrogate from biasing the future search. intuitively this threshold value could be set to the fitness of $Baseline$. Additionally, the effect of varying the initial population size $p_0$ on convergence could be studied, and compared to a non-surrogate counterpart approach. 

Currently, the number of surrogate levels $b$ is a hyper-parameter of our framework. We are considering to retire $b$ following an adaptive approach that starts with a fixed starting sample size, and progresses from one surrogate level to the next as long as we continue to observe improvements on best individual fitness measured by the original function. 

Additionally, we are considering strategies to influence the selection and mutation operators by our knowledge of the coexistence of solutions evaluated using the original function among a majority of solutions evaluated using surrogates. 

A very interesting extension of this work would be to evaluate our framework on other Evolutionary Computation $(EC)$ approaches. Recent methods using Particle Swarm Optimization $(PSO)$ are showing promise in solving large scale feature selection tasks\cite{pso1, pso2}. This could prove complementary given our framework demonstrated scalibility for large datasets in terms of number of instances.

\end{document}